\documentclass[pdflatex,sn-mathphys-num]{sn-jnl}

\usepackage{amsmath,amssymb,amsfonts}%
\usepackage{amsthm}%
\usepackage[title]{appendix}%
\usepackage{xcolor}%
\usepackage{textcomp}%
\usepackage{manyfoot}%
\usepackage{booktabs}%
\usepackage{array}
\usepackage{algorithm}%
\usepackage{algorithmicx}%
\usepackage{algpseudocode}%
\usepackage{listings}%
\usepackage{graphicx}  
\usepackage{multirow}  
\usepackage{adjustbox} 
\usepackage{bbding}

\usepackage{amsthm}
\newtheorem{corollary}{Corollary}

\usepackage{subfig}

\theoremstyle{thmstyleone}%
\newtheorem{theorem}{Theorem}
\newtheorem{proposition}[theorem]{Proposition}%

\theoremstyle{thmstyletwo}%

\theoremstyle{thmstylethree}%

\begin{document}


\title[Article Title]{LSAP: Rethinking Inversion Fidelity, Perception and Editability in GAN Latent Space}

\author[1]{\fnm{Xuekun} \sur{Zhao}}\email{zhaoxuekun@bupt.edu.cn}
\author[1]{\fnm{Pu} \sur{Cao}}\email{caopu@bupt.edu.cn}
\author[1]{\fnm{Xiaoya} \sur{Yang}}\email{yangxiaoya@bupt.edu.cn}
\author[1]{\fnm{Mingjian} \sur{Zhang}}\email{zhangmingjian2024@bupt.edu.cn}
\author[1]{\fnm{Lu} \sur{Yang}}\email{soeaver@bupt.edu.cn}
\author*[1]{\fnm{Qing} \sur{Song}}\email{priv@bupt.edu.cn}

\affil[1]{Beijing University of Posts and Telecommunications}

\abstract{
As research on image inversion advances, the process is generally divided into two stages. The first step is Image Embedding, involves using an encoder or optimization procedure to embed an image and obtain its corresponding latent code. The second stage, referred to as Result Refinement, further improves the inversion and editing outcomes. Although this refinement stage substantially enhances reconstruction fidelity, perception and editability remain largely unchanged and are highly dependent on the latent codes derived from the first stage. Therefore, a key challenge lies in obtaining latent codes that preserve reconstruction fidelity while simultaneously improving perception and editability. In this work, we first reveal that these two properties are closely related to the degree of alignment (or disalignment) between the inverted latent codes and the synthetic distribution. Based on this insight, we propose the \textbf{ Latent Space Alignment Inversion Paradigm (LSAP)}, which integrates both an evaluation metric and a unified inversion solution. Specifically, we introduce the \textbf{Normalized Style Space ($\mathcal{S^N}$ space)} and \textbf{Normalized Style Space Cosine Distance (NSCD)} to quantify the disalignment of inversion methods. Moreover, our paradigm can be optimized for both encoder-based and optimization-based embeddings, providing a consistent alignment framework. Extensive experiments across various domains demonstrate that NSCD effectively captures perceptual and editable characteristics, and that our alignment paradigm achieves state-of-the-art performance in both stages of inversion. 
}

\keywords{
GAN Inversion, Latent Space, Style Space, GAN 
}

\maketitle

\section{Introduction}
\label{sec1}

In recent years, Generative Adversarial Networks (GANs) \cite{goodfellow2014generative} have been widely applied to various vision tasks \cite{ledig2017photo,yang2021attacks}, greatly advancing the capability of image synthesis. Style-based generative models \cite{karras2019style,karras2020analyzing,karras2021alias} have further improved both the realism and resolution of generated images, achieving state-of-the-art performance. The intermediate latent space $\mathcal{W}$ space in StyleGAN encodes high-level semantic information. Leveraging this strong prior, a well-trained generator has demonstrated remarkable power and has significantly enhanced multiple downstream tasks compared with traditional approaches, such as neural talking head \cite{prajwal2020lip,yin2022styleheat}, face parsing \cite{yang2021quality,zhang2021datasetgan}, and style transfer \cite{li2020accelerate,yang2022pastiche}.

These applications rely on latent codes that are inherently accessible for synthetic images but not directly applicable to real ones. To address this issue, inversion methods have been developed to embed real images into the latent space of GANs through various approaches. Existing studies can be broadly categorized into two stages. The first stage, termed Image Embedding, focuses on obtaining latent codes, typically by training an encoder or optimizing the reconstruction loss. The second stage, researchers employ diversiform strategies to improve inversion and editing results, e.g., predicting generator weights \cite{alaluf2022hyperstyle, dinh2022hyperinverter}, predicting intermediate feature \cite{wang2022high, parmar2022spatially}, and finetuning the generator \cite{roich2021pivotal, feng2022near}, which we named Result Refinement. Previous works \cite{tov2021designing} illustrate that fidelity, perception, and editability are three essential attributes of successful inversion. However, most refinement approaches mainly emphasize improving fidelity, such as preserving visual details (e.g., background, hats, or eyeglasses), while the perception and editability of the results remain largely dependent on the latent codes obtained in the first stage. Therefore, to achieve a better balance among these three aspects, a more robust and perceptually aligned latent code embedding technique is still required.

The latent space obtained through random sampling and transformation follows a particular distribution, which we refer to as the synthetic distribution. Intuitively, latent codes drawn from this distribution exhibit superior performance. Supervision from the discriminator constrains the sampled latent codes to generate photorealistic images. Furthermore, editing directions can be derived through sampling \cite{shen2020interfacegan} and analysis \cite{harkonen2020ganspace} within the synthetic latent space. Consequently, the key to achieving high perception and editability lies in ensuring alignment between the inverted latent codes and the synthetic distribution. An existing method \cite{tov2021designing} addresses this issue by employing a latent code discriminator, thereby achieving more reasonable perception and editability. However, two major limitations remain. First, it limits the reconstruction performance since introducing a discriminator makes training unstable. Second, this method is inherently incompatible with optimization-based inversion frameworks. Therefore, our primary motivation is to develop an alignment paradigm that bridges the embedding latent space and the synthetic latent space, applicable to both encoder-based and optimization-based inversion methods, while preserving strong reconstruction capability.

\begin{figure*}[htbp]
	\begin{center}
		\includegraphics[width=1.0 \linewidth]{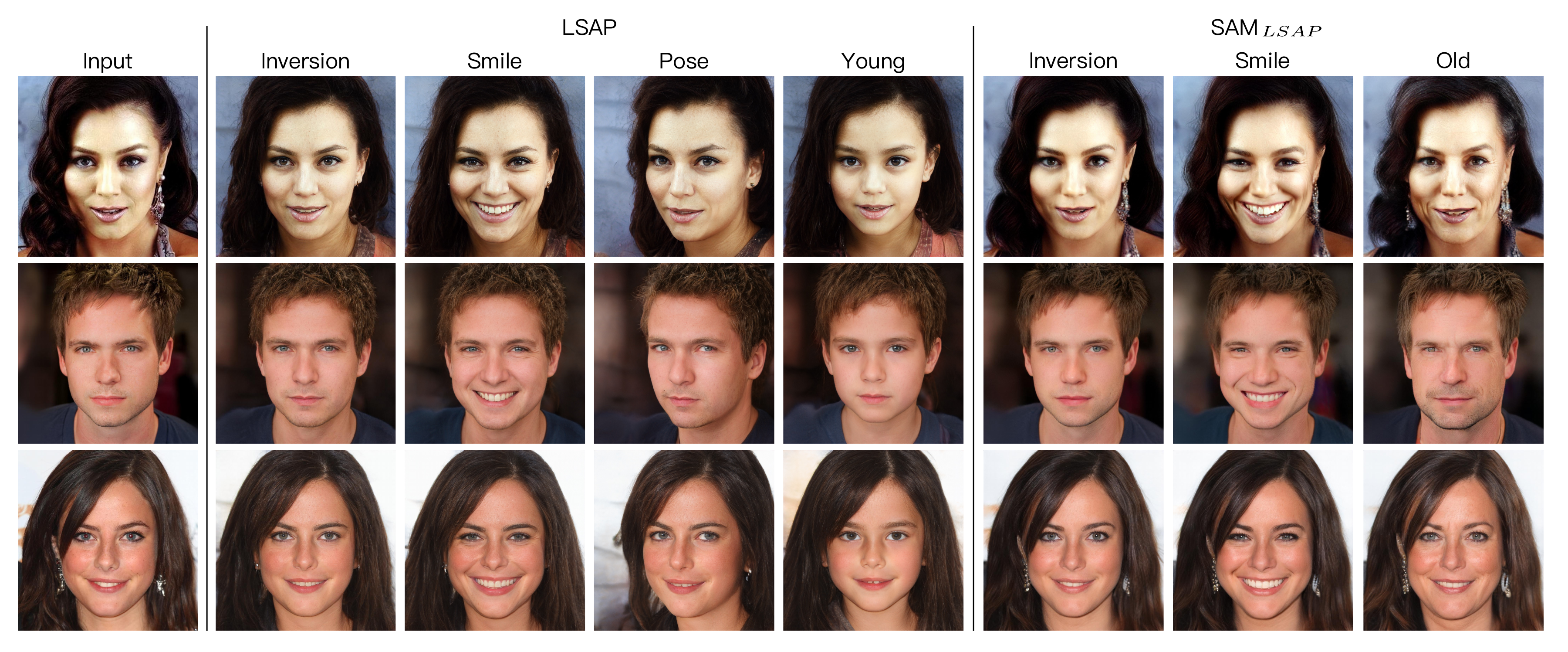}
	\end{center}
	\caption{\textbf{Inversion and editing results produced by LSAP and SAM$_{LSAP}$ \cite{parmar2022spatially}}. Our method enhances image quality and editability while preserving reconstruction fidelity. It is compatible with the two-stage inversion framework and achieves better performance.}
	\label{fig:introduce}
\end{figure*}

In this work, we conduct a comprehensive analysis of the disalignment problem in GAN inversion and propose the Latent Space Alignment Inversion Paradigm (LSAP). Specifically, we first introduce the Normalized Style Space ($\mathcal{S^N}$ space) and demonstrate that it provides a more suitable and efficient representation for measuring disalignment than the conventional $\mathcal{Z}/\mathcal{W}/\mathcal{S}$ space. We further define a novel metric, the Normalized Style Space Cosine Distance (NSCD), to quantitatively evaluate inversion methods at the latent code level, which has been experimentally shown to correlate strongly with perception and editability. Building on these insights, we incorporate our alignment solution into both encoder-based and optimization-based inversion frameworks by employing an alignment loss derived from NSCD. Extensive experiments validate the effectiveness and generality of our proposed paradigm.
Our approach achieves the best trade-off among fidelity, perception, and editability in encoder-based methods, and significantly enhances perception and editability in optimization-based methods. Moreover, LSAP attains state-of-the-art performance when integrated with existing refinement frameworks such as HFGI \cite{wang2022high}, SAM \cite{parmar2022spatially}, and PTI \cite{roich2021pivotal}, further demonstrating its versatility and potential. As shown in Figure~\ref{fig:introduce}, our visual results exhibit natural and faithful reconstruction quality.

The key contributions of this work are summarized as follows:
\begin{itemize}
\item We revisit the concepts of fidelity, perception, and editability in the inversion task. By dividing the inversion process into two stages—Image Embedding and Result Refinement—we observe that fidelity is primarily enhanced in the second stage, while perception and editability are closely related to the alignment between the inverted latent codes and the synthetic distribution.
\item We propose an effective and generalizable Latent Space Alignment Inversion Paradigm(LSAP) that provides both a measurement metric and alignment solutions to improve perception and editability. 
\item To demonstrate the effect of our aligning paradigm, we conduct extensive experiments across various domains. The proposed Normalized Style Space Cosine Distance (NSCD) quantitatively reflects perception and editability in a numerical manner. Our alignment paradigm reaches better trade-offs between fidelity and perception as well as fidelity and editability. When applied to Result Refinement methods, LSAP$_E$ achieves state-of-the-art performance.
\end{itemize}

\section{Related Work}
\subsection{GAN Inversion} As discussed above, the inversion process can be divided into two main stages. In the first stage, an initial latent code is obtained either through optimization or by using an encoder. Optimization-based approaches\cite{karras2020analyzing, creswell2018inverting, abdal2019image2stylegan, abdal2020image2stylegan++} typically achieve higher reconstruction fidelity but are computationally expensive, often requiring several minutes per image. In contrast, encoder-based methods \cite{tov2021designing, richardson2021encoding, wei2022e2style, guan2020collaborative, creswell2018inverting} provide much faster inference but generally yield inferior reconstruction quality. The second stage focuses on refining the inversion and editing results using various strategies. Some methods \cite{alaluf2022hyperstyle, dinh2022hyperinverter} adjust the generator’s convolutional weights via a hypernetwork \cite{ha2016hypernetworks}. ReStyle \cite{alaluf2021restyle} introduces an iterative refinement mechanism, progressively updating the latent code using a residual-based encoder. HFGI \cite{wang2022high} proposes a distortion consultation approach for high-fidelity reconstruction. SAM \cite{parmar2022spatially} predicts the invertibility of different image segments to map them into corresponding intermediate layers.Finally, generator tuning methods \cite{roich2021pivotal, feng2022near} achieve the highest inversion accuracy but are extremely time-consuming.

\subsection{GAN-based Manipulation}
Owing to the rich semantic information embedded in the latent space of GANs \cite{karras2019style, karras2020analyzing, karras2021alias}, numerous studies have proposed methods to control generated images by manipulating their latent representations. Some methods \cite{denton2019detecting, goetschalckx2019ganalyze, spingarn2020gan, shen2020interfacegan} discover attribute editing directions (e.g., smile, gender, age, pose) using semantic supervision or annotated labels. Others explore meaningful manipulation directions in an unsupervised \cite{harkonen2020ganspace, shen2021closed, voynov2020unsupervised, wang2021geometry} or self-supervised \cite{jahanian2019steerability, plumerault2020controlling} manner. In addition, language–image models have been utilized to perform text-guided image editing by back-propagating gradients from textual objectives \cite{patashnik2021styleclip}. Some works \cite{sun2021secgan,kim2022style} further integrate segmentation information to improve editing precision and visual consistency, and this concept can potentially be extended to the human body domain via body GANs \cite{fruhstuck2022insetgan,fu2022stylegan} and human parsing techniques \cite{yang2019parsing,yang2020renovating,yang2021quality,yang2022part} in the future. Since most of these manipulation approaches rely directly on latent code representations, editability has become a crucial characteristic in evaluating inversion performance.

\section{Latent Space Disalignment}
\label{sec:disalignment}
In this section, we first revisit the origins of fidelity, perception, and editability, and highlight that the latter two are strongly influenced by the alignment (or disalignment) between the inverted latent codes and the synthetic distribution. To better illustrate and address this issue in the inversion task, we formally define and quantify the degree of disalignment in the inversion process.
\subsection{Fidelity, Perception and editability}
\label{sec:inversion_ability}
As first introduced by Tov \cite{tov2021designing}, fidelity\footnote{ \emph{Image distortion} is originally used in e4e \cite{tov2021designing} . To represents the ability of inversion methods, we use fidelity instead of it.}, perception and editability are three fundamental characteristics of GAN inversion methods. Fidelity measures the reconstruction capability, requiring the inversion process to embed an image into the latent space such that it can be faithfully reconstructed. Perception evaluates the perceptual quality of reconstructed images, typically reflecting attributes such as sharpness and naturalness. Finally, editability represents the degree to which the inverted latent codes can be manipulated, serving as a comprehensive indicator that encompasses editing effectiveness, attribute disentanglement, and related editing capabilities.

\noindent\textbf{Source} We first trace these three characteristics.  Minimizing image distortion is a fundamental objective in nearly all inversion methods, granting the algorithm the ability to faithfully reconstruct input images. Perception arises from the powerful generative capacity of GANs, whose generator is trained under the supervision of a discriminator to produce high-resolution, photorealistic results. Editability, in turn, benefits from the semantically rich latent space of GANs—given a specific editing direction, the latent codes can be modified to alter corresponding attributes. However, both perception and editability are conditional properties. Under the discriminator’s supervision, the latent space of a trained GAN is constrained to fit the dataset distribution, ensuring that latent codes sampled from this space generate high-quality, realistic images. In contrast, out-of-distribution latent codes may lead to degraded or unrealistic results—an effect also observed in latent code truncation, where latent codes closer to the mean vector tend to yield higher-quality generations. Furthermore, editing directions are typically derived by sampling latent codes \cite{shen2020interfacegan} or analyzing generator weights \cite{shen2021closed}, both of which inherently depend on a specific latent space structure within the GAN. We refer to this underlying latent space distribution as the synthetic distribution, which is transformed by the pre-trained generator from a multivariate standard normal distribution.
We name the latent space distribution in GAN as synthetic distribution, which is converted by pre-trained networks from multivariate normal standard distribution. 

\noindent\textbf{Impacts from the Two Inversion Stages} The inversion process can be divided into two stages: Image Embedding and Result Refinement. In the first stage, latent codes are obtained either through an encoder or by optimization that minimizes image distortion. At this point, the reconstruction error remains relatively high.
In the Result Refinement stage, methods aim to recover finer visual details (e.g., background, clothing) by adjusting the generator’s weights \cite{alaluf2022hyperstyle} or intermediate features \cite{wang2022high}. This stage further enhances fidelity and can even invert out-of-distribution images \cite{roich2021pivotal,feng2022near}. However, perception and editability are largely inherited from the latent codes produced in the first stage. In practice, if the initial latent codes lack editability or fail to yield perceptually realistic images, the refined results will still exhibit the same limitations. Therefore, a key challenge lies in obtaining latent codes with superior perceptual and editable qualities. In this work, we primarily focus on the Image Embedding stage to investigate how fidelity, perception, and editability emerge from the latent codes.

\subsection{Disalignment Formulation}
\label{sec:disalignment_formulation}

To illustrate the disalignment between the synthetic and inverted latent spaces, we first define a reference latent space $\mathcal{P}$, denoting the inverse latent space as $\mathcal{P}_{inv}$ and the synthetic latent space as $\mathcal{P}_{syn}$. Let $G_{\mathcal{P}}$ represent the generator that maps from $\mathcal{P}$ space to the image space. Suppose that $\mathcal{Z}$ follows a multivariate standard normal distribution, and $\mathcal{X}$ denotes the real image distribution. We define two mapping functions $F: \mathcal{Z}\to\mathcal{P}_{syn}$ and $I:\mathcal{X}\to\mathcal{P}_{inv}$. In practice, $I$ serves as an embedding function, mapping real images into the latent space $\mathcal{P}$. Meanwhile, $F$ represents a mapping function composed of several layers preceding the generator in the GAN architecture. It is important to note that alignment does not imply that the inverse distribution should be identical to the synthetic one. Instead, the goal is to ensure that the inverted latent codes lie within the high-probability regions of the synthetic distribution. Accordingly, we define the degree of disalignment $\mathcal{D}$ between these two spaces as follows:
\begin{align}
\label{equation:disalignment}
\mathcal{D}=-\mathbb{E}_{x_{inv}\sim\mathcal{P}_{inv}}p_{syn}(x_{inv})
\end{align}

\begin{figure}[htbp]
	\centering
	\includegraphics[width=1.0 \linewidth]{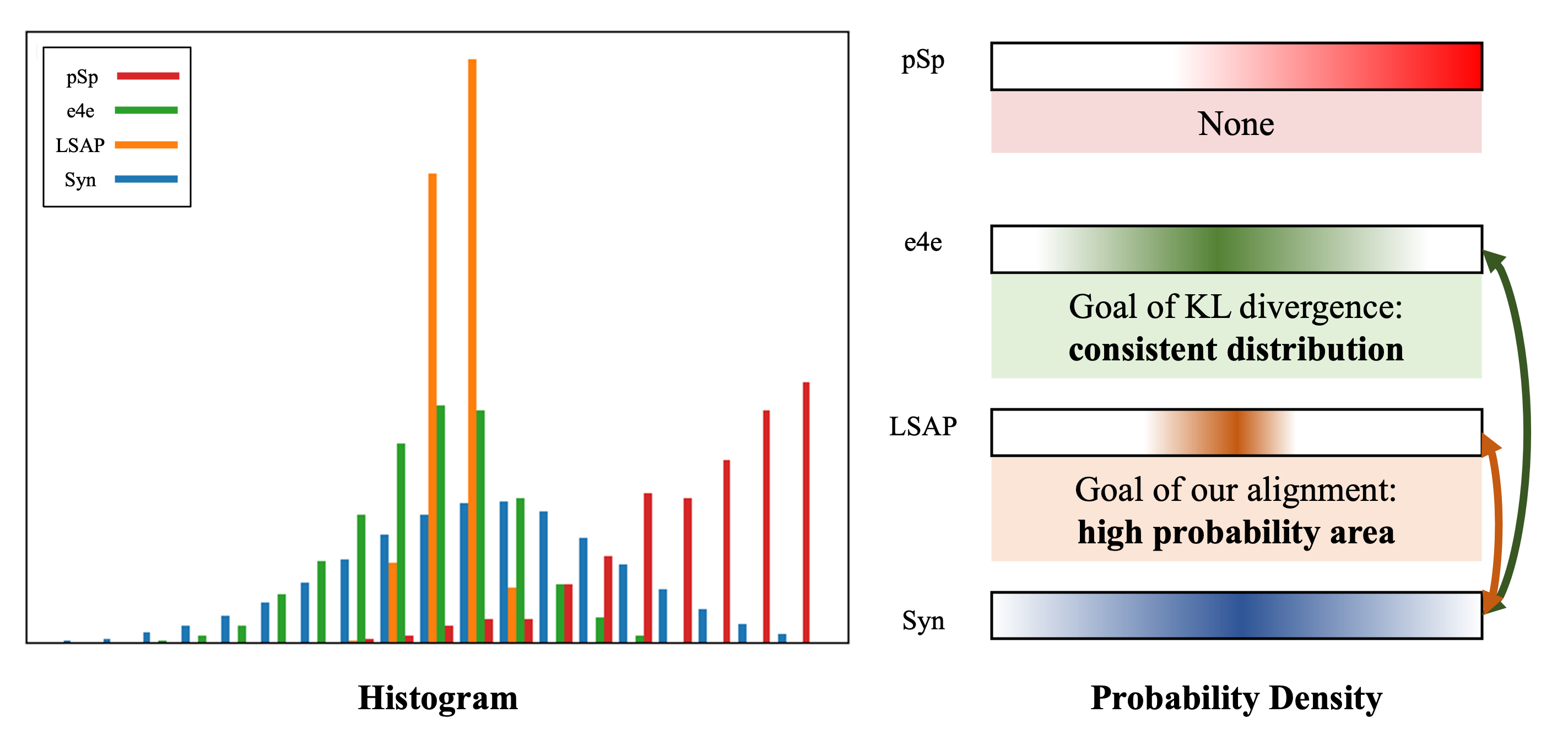}
	\caption{\textbf{Illustration of Latent Space Distributions.} We invert all images from the CelebA-HQ test split into the latent space and visualize their distribution in the $\mathcal{S}^\mathcal{N}$ space. Our alignment solution ensures that the embedded latent codes are located in the high-probability regions of the synthetic distribution, thereby preserving both perception and editability.}
\label{fig:distribution}
\end{figure}

\noindent\textbf{Compared to Kullback-Leibler Divergence} Another potential way to measure and optimize disalignment is through the Kullback–Leibler (KL) divergence. In practice, the latent code discriminator used in e4e \cite{tov2021designing} can be interpreted as an implicit attempt to minimize $D_{KL}(\mathcal{P}_{syn}||\mathcal{P}_{inv})$. However, as discussed earlier, the objective of alignment is not to make the inverse and synthetic distributions identical, but rather to ensure that inverted latent codes reside in the high-probability regions of the synthetic distribution. Moreover, KL divergence can not be directly measured in inversion task. Consequently, it cannot be applied to optimization-based inversion methods, nor does it serve as an effective metric for evaluating the characteristics of inversion methods. This conceptual difference is also illustrated in Figure~\ref{fig:distribution}.

In Figure~\ref{equation:disalignment}, two vital parts of disalignment measurement are which latent space is adequate to measure and how to measure $p_{syn}(x_{inv})$ for given sample. We will respectively answer these two questions in the following parts.

\section{Latent Space Alignment Inversion Paradigm}
In this section, we present the Latent Space Alignment Inversion Paradigm (LSAP) which is designed to both measure and enhance the perception and editability of inversion methods.
Specifically, we introduce a novel latent representation, the Normalized Style Space ($\mathcal{S^N}$) and propose the Normalized Style Space Cosine Distance (NSCD) as a measurement. Furthermore, we develop a set of generalized alignment solutions for the Image Embedding stage,
including LSAP$_E$ for encoder-based methods and LSAP$_O$ for optimization-based methods.

\subsection{Normalized Style Space}
Although the $\mathcal{Z}/\mathcal{W}/\mathcal{W}^+$ spaces have been predominantly used in prior research, we introduce a new latent representation, the Normalized Style Space ($\mathcal{S^N}$), and demonstrate that it provides a more effective basis for measuring disalignment.

To better motivate our formulation, we first revisit the existing latent spaces. Given a random latent variable $z$ sampled from the $\mathcal{Z}$ space, the mapping network transforms it into $w$ in the $\mathcal{W}$ space. Subsequently, affine transformation modules are applied to $w$ at each resolution level, producing a set of style parameters is $s=\{s_1, s_2, \dots, s_k\}$, where each style vector is computed as $s_i=A_iw+b_i$. The space spanned by these style vectors is referred to as the ($\mathcal{S}$ space). 

\begin{proposition}
\label{property:s_scale_independent}
Suppose that $s=\{s_1, s_2, \dots, s_k\}$ is a set of $\mathcal{S}$ space latent codes and corresponding to image $x=G_\mathcal{S}(s)$. For $\forall a \in \mathbb{R}^+$ and $\forall l \in \{1, \cdots, k\}$, if $s^\prime=\{s^\prime_1, s^\prime_2, \dots, s^\prime_k\}$ follows:
$$ s^\prime_i=
\begin{cases}
s_i, & i\neq l \\
a\times s_i, &i=l  
\end{cases}
$$
we have $x=G_\mathcal{S}(s)=G_\mathcal{S}(s^\prime)$.

\begin{proof}
According to StyleGAN2, style latent codes are applied through a weight demodulation mechanism. For $l$th convolutional layer, the kernel weights $W_{i,j,k}$ are modulated by the corresponding style latent code $s_l$ as follows:
\begin{align}
\label{equation:conv_scale}
W^{\prime}_{ijk}=s_l^i \times W_{ijk},
\end{align}
where $i,j,k$ index the input feature maps, output feature maps, and the spatial kernel footprint, respectively.

To incorporate instance normalization within the convolution operation, StyleGAN2 demodulates each output feature map by a normalization factor $\sigma_j=\sqrt{\sum_{i,k}{W^{\prime}_{ijk}}^2}$, under the assumption that the input activations are i.i.d. random variables with unit standard deviation (ignoring the small numerical constant $\epsilon$ used for stability):

\begin{align}
\label{equation:conv_demoluate1}
W_{i j k}^{\prime \prime}=\frac{W_{i j k}^{\prime}}{ \sqrt{\sum_{i, k} {W_{i j k}^{\prime}}^{2}}}
\end{align}
By substituting Equation \ref{equation:conv_scale} into Equation \ref{equation:conv_demoluate1}, we obtain:
\begin{align}
\label{conv_demodulate2}
W_{i j k}^{\prime \prime}=\frac{s_l^i\times W_{i j k}}{ \sqrt{\sum_{i, k} (s_l^i \times W_{i j k}})^{2}}
\end{align}
Suppose that $\hat{s_l}=a\times s_l$, 

\begin{align}
\label{conv_demodulate3}
\hat{W}_{i j k}^{\prime \prime}&=\frac{\hat{s}_l^{i}\times W_{i j k}}{ \sqrt{\sum_{i, k} (\hat{s}_l \times W_{i j k})^{2}}} \nonumber \\
&=\frac{a \times s_l^i\times W_{i j k}}{ \sqrt{\sum_{i, k} (a \times s_l^i \times W_{i j k})^{2}}} \nonumber \\
&=\frac{s_l^i\times W_{i j k}}{ \sqrt{\sum_{i, k} (s_l^i \times W_{i j k})^{2}}}=W_{i j k}^{\prime \prime}
\end{align}
Thus, if scale $s$ by $a\in\mathbb{R}^+$ in an arbitrary layer, convolution weights are identical, meaning generated images are the same.
\end{proof}

\end{proposition}

Property~\ref{property:s_scale_independent} illustrates that the Style Space($\mathcal{S}$ space) latent codes are scaled-independent in every component. When these codes are projected onto the unit hypersphere, codes sharing the same angular direction produce identical outputs. Leveraging this property, we construct a new latent representation, the Normalized Style Space ($\mathcal{S^N}$), in which the style codes from $\mathcal{S}$ space are normalized by their Euclidean norm. Formally, this can be expressed as:
\begin{align}
\label{equation:sn}
s^N_i=\frac{s_i}{\Vert s_i\Vert_2}=\frac{A_iw+b_i}{\Vert A_iw+b_i \Vert_2}
\end{align}

To demonstrate the differences among various latent spaces in measuring disalignment, we conduct extensive analyses:
\begin{proposition}
    \label{property:m2o_s}
    Given a sets of $\mathcal{S}$ space latent codes $s=\{s_1,\dots, s_k\}\neq \textbf{0}$, $\exists s^\prime=\{s_1^\prime,\dots, s_k^\prime\} \neq s$ such that $G_\mathcal{S}(s)=G_\mathcal{S}(s^\prime)$. 
    \begin{proof}
        According to Property~\ref{property:s_scale_independent}, for $\forall l \in \{1, \cdots, k\}$ when $s_l^\prime=a\times s (a\in\mathbb{R}^+)$ and $s_i^\prime=s_i$($i\neq l$), we have $G_\mathcal{S}(s)=G_\mathcal{S}(s^\prime)$. Since $s_l \neq \textbf{0}$, $s_l^\prime\neq s_l$.
    \end{proof}
\end{proposition}

\begin{proposition}
    \label{property:m2o_wz}
    For $l$th layer ($\forall l \in \{1, \cdots, k\}$), define $F_l:\mathcal{Z/W}\to\mathcal{S}$ as the mapping function between $\mathcal{S}$ and $\mathcal{Z}/\mathcal{W}$ space. For all $p_l \in\mathcal{Z/W}$ ($F_l(p)\neq \textbf{0}$), exist $p_l'\neq p_l$ such that the corresponding $S$ space latent codes satisfy: $s_l^\prime=a\times s_l$ ($a \in \mathbb{R}^+$), where $s_l=F_l(p_l)$ and $s_l^\prime=F_l(p_l^\prime)$.
\begin{proof} We prove this property separately under $\mathcal{W}$ and $\mathcal{Z}$ spaces. Since cases under each layer level are the same without loss of generality, to express concisely, we consider the situation under an arbitrary layer and ignore $l$ in the later formulation.

    \noindent$\mathcal{W}$ space The mapping function between $\mathcal{W}$ and $\mathcal{S}$ space is established by linear projection in generator, as follows:
    \begin{align}
    \label{equation:w2s}
        s=F(w)=Aw+b
    \end{align}
    If $\exists y$, such that 
    \begin{align}
    \label{equation:y}
        Ay=(a-1)b
    \end{align} and let 
    \begin{align}
        w^\prime=aw+y
    \end{align} we have 
    \begin{align}
        s^\prime=Aw^\prime+b=A(aw+y)+b=aAw+ab=as
    \end{align}
    
    In StyleGAN, $A\in\mathbb{R}^{m\times n}(m\leq n)$ may not be a square matrix in some resolution levels and the rank of $A$ is unstable. It indicates Equation~\ref{equation:y} can not be solved by inverse of $A$ directly. We can obtain $y$ by solving the least squares problem: 
    \begin{align}
        \label{equation:lse}
        \min_{y}\Vert Ay-(a-1)b\Vert
    \end{align} Hence, for $\forall w$, when $w^\prime=aw+y$, $F(w)=a\cdot F(w^\prime)$. In addition, we can prove $w^\prime\neq w$ by the counterfactual method. If $w^\prime=w$, we have $y=(1-a)w$ and $A(1-a)w=(a-1)b$, so $Aw=-b$ and $s=0$. Due to $s\neq 0$, $w^\prime\neq w$ and $w^\prime=aw+y$, $F(w)=\cdot F(w^\prime)$, we prove that property holds in $\mathcal{W}$ space.

    \noindent$\mathcal{Z}$ space Although we have proved in $\mathcal{W}$ space, the mapping function between $\mathcal{Z}$ and $\mathcal{W}$ or $\mathcal{Z}$ and $\mathcal{S}$ is represented by a multilayer perception, which is difficult to prove directly by formula. Fortunately, as the objective function is defined, we can obtain $z'$ by optimization, satisfying $s=F(z)=a\times F_(z')=ks'$ and $z'\neq z$.
\end{proof}

\end{proposition}

\begin{corollary}
    \label{corollary:m2o}
    Given a sets of latent codes $p=\{p_1,\dots, p_k\}$ in $\mathcal{Z}/\mathcal{W}/\mathcal{S}$ space and $p\neq 0$, $\exists p^\prime=\{p_1^\prime,\dots, p_k^\prime\} \neq p$ such that $G_\mathcal{P}(p)=G_\mathcal{P}(p^\prime)$. 
\end{corollary}

According to Corollary~\ref{corollary:m2o}, different latent codes in $\mathcal{Z}/\mathcal{W}/\mathcal{S}$ space can generate the same images, which implies the disalignment degree of these latent codes can not reflect discrepancies in generated results. Hence, we choose $\mathcal{S^N}$ as reference space to measure disalignment in inversion.

\subsection{Normalized Style Space Cosine Distance}
\label{sec:NSCD}
As illustrated in Figure~\ref{equation:disalignment}, the probability of inverted latent codes under the synthetic distribution needs to be estimated. However, it is intractable to compute $p_{syn}(x_{inv})$ directly, since the analytical form of $p_{syn}$ is unknown. Inspired by the latent code truncation technique, we find that using distance between inverse code and mean code instead of $p_{syn}(x_{inv})$ is a simple but efficient way. Code near mean code has a high probability practically. When we denote $\mathcal{S^N}$ space as reference space, we can use cosine distance to measure disalignment and define NSCD as follow:
\begin{align}
\label{equation:cosine_distance}
    \mathrm{NSCD} &= 1-\mathbb{E}_{s_{inv}\sim\mathcal{S}_{inv}}[\cos(s^N_{inv}, \mu_{syn})]\nonumber \\ 
    &=1-\mathbb{E}_{s_{inv}\sim\mathcal{S}_{inv}}[s^N_{inv}\cdot\mu_{syn}^T]
\end{align}
Notably, since $s^N_{inv}$ represents the inverted latent code, the cosine distance is differentiable and can therefore be minimized during the inversion process.
The small value of NSCD means that $\mathcal{S}^\mathcal{N}_{inv}$ space aligns with $\mathcal{S}^\mathcal{N}_{syn}$ space. Moreover, NSCD effectively reflects the perceptual quality and editability of reconstructed images, as will be demonstrated in our qualitative and quantitative experiments.

\subsection{Alignment Inversion}

\label{sec:lsap}
In the Image Embedding phase, inversion methods aim to embed images into the latent space through either an encoder-based or an optimization-based approach. The overall process can be summarized as follows:
\begin{align}
    \label{equation:inv_split}
    p^*&=\underset{p}{\arg\min} [\mathcal{L}(x, G_{\mathcal{P}}(p)] \\
    E^*&=\underset{E}{\arg\min} [\mathbb{E}_{x\sim\mathcal{X}}(\mathcal{L}(x, G_{\mathcal{P}}(E(x))))]
\end{align}
Here, $x$ denotes the input image, $\mathcal{X}$ represents the image dataset, $\mathcal{L}$  is the image-level loss function (e.g., MSE, LPIPS \cite{zhang2018unreasonable}, identity loss \cite{deng2019arcface}) and $E$ denotes the encoder.  Since inversion methods are primarily supervised at the image level, they lack explicit constraints on the distribution of inverted latent codes. To develop a unified solution for training the encoder or optimizing latent codes, we constrain the degree of disalignment by introducing an additional alignment term into the overall loss function  $\mathcal{L}$.

\begin{figure*}[htbp]
  \centering
	\subfloat[LSAP$_E$]{
		\centering
		\includegraphics[width=0.48\textwidth]{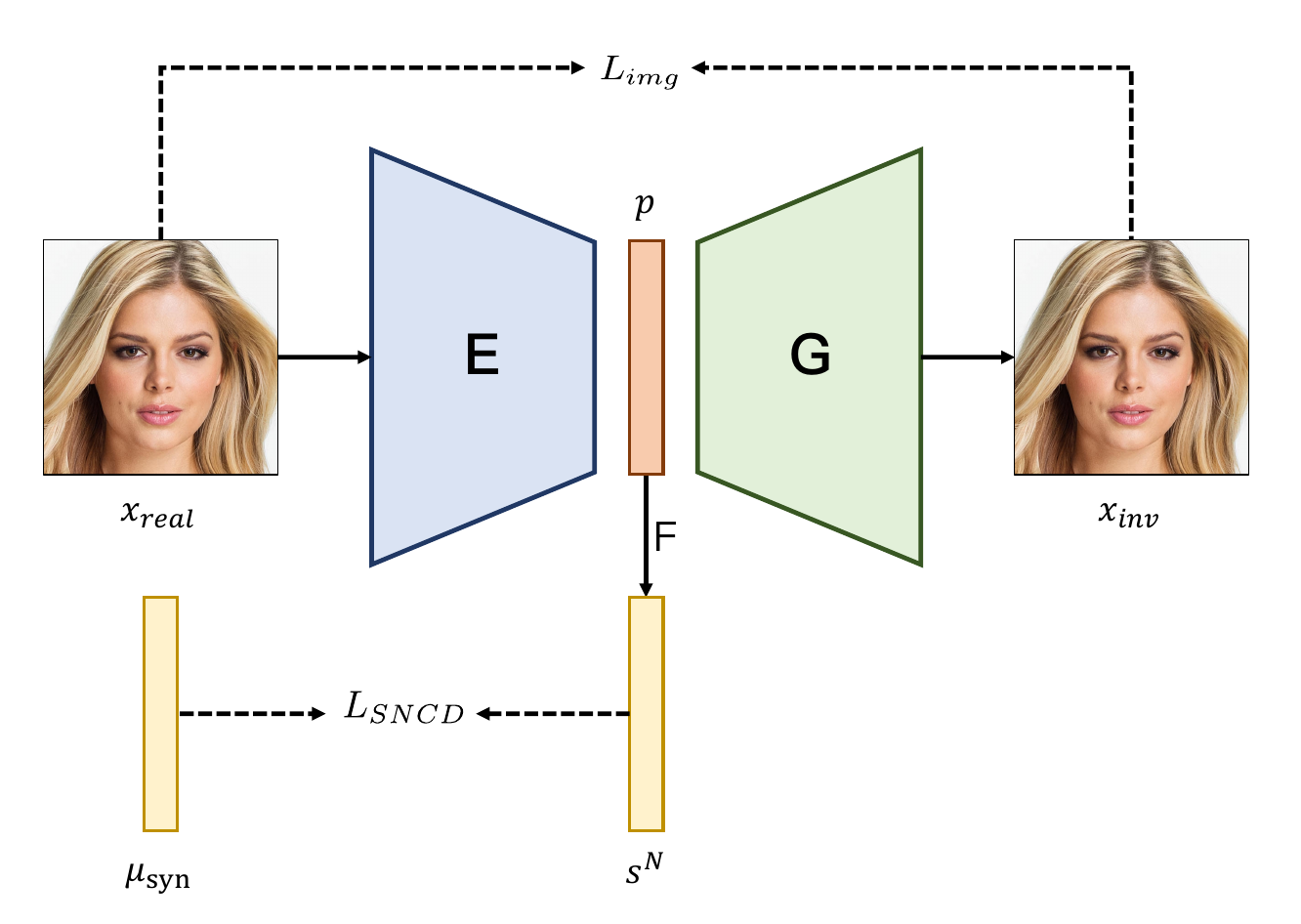}
      	\label{fig:lsap_enc}
	}
	\subfloat[LSAP$_O$]{
		\centering
		\includegraphics[width=0.48\textwidth]{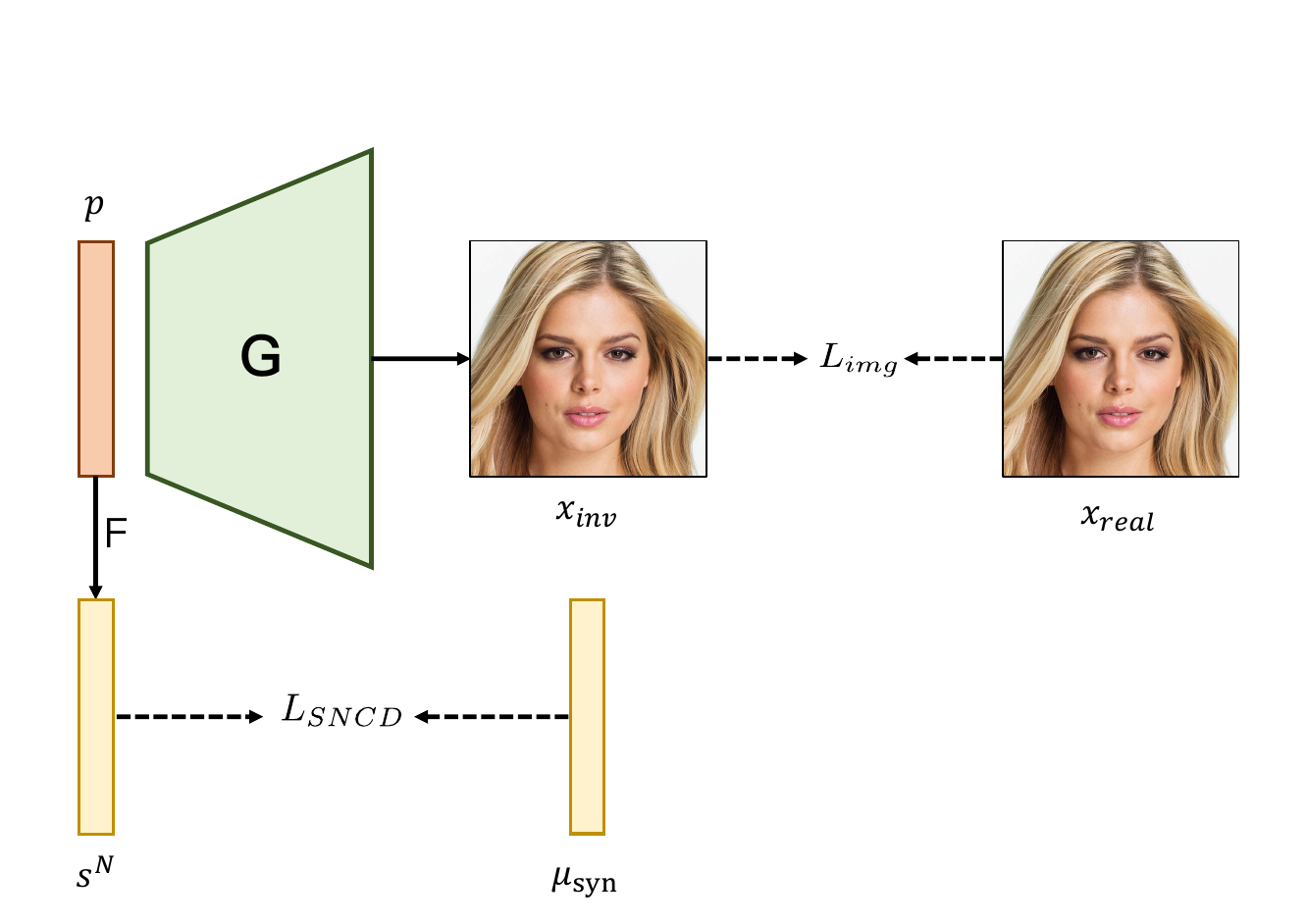}
		\label{fig:lsap_opt}
}
  \caption{\textbf{Alignment inversion solutions of LSAP.} We show the details of encoder-based and optimization-based inversion methods in our alignment paradigm. The pivotal part is the $L_{NSCD}$, which represents the disalignment degree of inverse latent codes.}
  \label{fig:lsap}
\end{figure*}

Benefiting from the differentiable property of NSCD, we incorporate it into the inversion framework to construct a direct and efficient alignment solution, as illustrated in Figure~\ref{fig:lsap}. According to Equation~\ref{equation:cosine_distance}, we first sample $k$ latent codes ($k=50,000$ in our experiments) from a multivariate normal distribution, and transform them into $\mathcal{S^N}$ using the pre-trained generator to obtain the mean latent code $\mu_{syn}$. Based on this, we define the alignment loss as follows:
\begin{align}
    \label{equation:loss_NSCD1}
    \mathcal{L}_{NSCD}(x)=1-(F(I(x))\cdot\mu_{syn}^T
\end{align}
where $I$ denotes the Image Embedding method. The NSCD loss
$L_{NSCD}$ is calculated by given images $x$ (i.e., a batch of images in encoder training or one image in optimization) in each iteration. In the following sections, we present the implementation details of our encoder-based and optimization-based methods, respectively.

\noindent\textbf{Encoder} The overall pipeline of the encoder-based alignment inversion method is illustrated in Figure~\ref{fig:lsap_enc}. Given real images, the encoder is trained by minimizing a combination of image-level and latent-level loss functions. Following \cite{tov2021designing, richardson2021encoding}, the image-level loss $L_{img}$ comprises a distortion loss, perceptual loss, and identity loss. In addition, the delta-regularization loss \cite{tov2021designing} is applied to the inverted codes to minimize the deviation among the $\mathcal{W}^+$ codes across different layers.
The overall training objective is defined as follows:
\begin{align}
\label{equation:encoder_loss}
\mathcal{L}=\mathcal{L}_2+\lambda_1\mathcal{L}_{lpips}+\lambda_2\mathcal{L}_{sim}+\lambda_3\mathcal{L}_{d-reg}+\lambda\mathcal{L}_{NSCD}
\end{align}
where $\lambda_1, \lambda_2, \lambda_3, \lambda$ are hyper-parameters to adjust the weight of each component in loss function. In encoder-based method, $L_{NSCD}$ aims to align the encoder's output space with synthetic latent space.

\noindent\textbf{Optimization} The overall pipeline of the optimization-based alignment inversion method is illustrated in Figure~\ref{fig:lsap_opt}. Unlike the encoder-based approach, the optimization-based inversion method updates the latent code iteratively. In this framework, the NSCD loss is employed to minimize the distance between the current latent code and the synthetic latent space. Following \cite{karras2020analyzing}, we adopt two types of loss functions, applied at the image level and the latent code level, respectively:
\begin{align}
\label{equation:optim_loss}
\mathcal{L}=\mathcal{L}_{lpips}+\lambda\mathcal{L}_{NSCD}
\end{align}
The encoder-based and optimization-based method are denoted as LSAP$_E$ and LSAP$_O$ respectively.

\section{Experiments}

In this section, we conduct extensive experiments to evaluate the effectiveness of LSAP across various domains, including face, object (cars), scene (churches), and animal (wild animals). The detailed implementation settings and experimental configurations are provided below.

\subsection{Implementation Details}
\label{sec:appendix_implementation_details}

\noindent\textbf{Datasets} We conduct experiments across four domains: faces, cars, churches, and wild animals, corresponding to the categories of human, object, scene, and animal, respectively. In all domains, we utilize the official StyleGAN2 generator as the pretrained model. For the face domain, we train LSAP$_E$ on the FFHQ dataset \cite{karras2019style} (70,000 images) and evaluate it on the CelebA-HQ \cite{liu2015deep,karras2017progressive} (2824 images). Editing directions obtained following \cite{shen2020interfacegan}. For car domain, we use Stanford Cars dataset \cite{krause20133d} which contains 8,144 training images, and we randomly select 1,000 images for evaluation. Image editing is performed following\cite{harkonen2020ganspace}. For the church domain, we adopt the LSUN Churches dataset \cite{yu2015lsun}, including 126,227 training images and 300 test images. For the wild animal domain, we employ the AFHQ-Wild dataset \cite{choi2020stargan} for both training and evaluation.

\noindent\textbf{LSAP$_E$} We set the input image resolution to $192\times 256$ for the car domain and $256 \times 256$ for all other domains. For data augmentation, we only apply random horizontal flipping. Model training uses the Ranger optimizer, which integrates Rectified Adam \cite{liu2019variance} and the Lookahead strategy \cite{zhang2019lookahead}, with a learning rate of $0.001$ . All experiments are conducted on a single GPU with a batch size of $8$, and we adopt the progressive training scheme from e4e  \cite{tov2021designing}. In LSAP$_E$, the perceptual loss weight $\lambda_1$ is set to $0.8$, the delta-regulation loss $\lambda_3$ is $2e-5$ , and the NSCD loss $\lambda$ is $0.5$ across all domains. The similarity loss weight  $\lambda_2$ is set to $0.1$ for the face domain using the pre-trained ArcFace model \cite{deng2019arcface} and to $0.5$ for other domains using MOCOv2  \cite{chen2020improved} and ResNet-50 \cite{he2016deep}. 

\noindent\textbf{Optimization-based Method} Following \cite{karras2020analyzing}, we employ the Adam optimizer \cite{kingma2014adam} to minimize the perceptual loss and NSCD loss, together with noise regularization. The loss weight $\lambda$ is set to $20$ for the $\mathcal{W^+}$ space and $5$ for the $\mathcal{W}$ space. 

\noindent\textbf{Result Refinement Method} We apply e4e and LSAP$_E$ to three Result Refinement methods---HFGI \cite{wang2022high}, SAM \cite{parmar2022spatially}, and PTI \cite{roich2021pivotal}--- to examine the impact of the Image Embedding step.  For HFGI, we use the official model weights to evaluate HFGI$_{e4e}$ and follow the released training script to train HFGI$_{LSAP}$,  where we simply replace the encoder weights of e4e with those of LSAP$_E$ while keeping the architecture unchanged. Since SAM only provides optimization code, we first obtain latent embeddings using each encoder and then optimize the latent codes with intermediate feature guidance for $500$ iterations, using a threshold of  $\tau=0.225$. For PTI, we take the inverse codes produced by the encoders as pivotal latents and fine-tune the generator for $350$ steps.

\noindent\textbf{Evaluation Pipeline} Since inversion and editing results are generated from different codebases, we perform all image-level evaluations on the saved image files. MSE, LPIPS, and identity similarity are computed at a resolution of $256\times256$ using the evaluation scripts from pSp\footnote{https://github.com/eladrich/pixel2style2pixel} \cite{richardson2021encoding}. For LEC and identity similarity, we apply adjusted editing factors to ensure consistent editing effects across all inversion methods, as shown in the qualitative results.

\noindent\textbf{Baselines} We conduct comprehensive comparisons across both stages of inversion. For Image Embedding methods, we compare our LSAP$_E$ with the encoder-based approaches pSp \cite{richardson2021encoding} and e4e \cite{tov2021designing}, and compare our optimization-based LSAP$_O$ with the projection method of StyleGAN2 \cite{karras2020analyzing}. Furthermore, we evaluate the performance of e4e and LSAP$_E$ under four Result Refinement methods: ReStyle \cite{alaluf2021restyle}, HFGI \cite{wang2022high}, SAM \cite{parmar2022spatially} and PTI \cite{roich2021pivotal}. Among these, HFGI and SAM operate as intermediate feature prediction methods, while PTI serves as a generator tuning method.

\noindent\textbf{Evaluation} We evaluate reconstruction fidelity using MSE and LPIPS \cite{zhang2018unreasonable} across all domains, and compute identity similarity for the face domain using \cite{deng2019arcface} between input and reconstructed images. To assess perception and editability, we adopt NSCD and latent editing consistency(LEC) \cite{tov2021designing}.  In addition, we measure identity preservation between the original images and their edited counterparts-under matched editing effects for each inversion method—to quantify identity robustness during manipulation.

\subsection{Quantitative Results}

We provide the reconstruction results in Table~\ref{table:face_inversion} to evaluate fidelity on face domain. Although the additional alignment loss leads to a slight reduction in fidelity compared with pSp and StyleGAN2 projection, it significantly improves perception and editability, as shown in the following analyses. Moreover, LSAP$_E$ consistently outperforms e4e. When applied to the Result Refinement methods, LSAP$_E$ surpasses e4e across all three. With ReStyle and HFGI, which use model to refine result and inference rapidly, LSAP$_E$ gains about 30\% improvement of MSE. PTI$_{LSAP}$ gains best results in inversion.

\begin{table*}[!h]
\centering
\resizebox{0.6\textwidth}{!}
{

\begin{tabular}{lccccc}
\toprule
\textbf{Method}                    &   \textbf{Type}  & \textbf{MSE} $\downarrow$     &      \textbf{Gain}      & \textbf{LPIPS} $\downarrow$               & \textbf{Similarity} $\uparrow$            \\ \midrule
pSp \cite{richardson2021encoding}                      & \emph{E}   & 0.0351 && 0.1628 & 0.5591 \\
e4e \cite{tov2021designing}                      & \emph{E}   & 0.0475 &                          & 0.1991                           & 0.4966                           \\
LSAP$_E$                  & \emph{E}   & 0.0397   &                       & 0.1766                           & 0.5305                           \\ \hline
StyleGAN2-$\mathcal{W}$ \cite{karras2020analyzing}  & \emph{O}   & 0.0696  &                         & 0.1987                           & 0.3066                           \\
LSAP$_O$-$\mathcal{W}$    & \emph{O}   & 0.0690  &                         & 0.1986                           & 0.2989                           \\
StyleGAN2-$\mathcal{W^+}$ \cite{karras2020analyzing} & \emph{O}   & 0.0279 && 0.1179 & 0.7463 \\
LSAP$_O$-$\mathcal{W^+}$  & \emph{O}   & 0.0359                           & & 0.1376                           & 0.6587                           \\ \hline
ReStyle$_{e4e}$ \cite{alaluf2021restyle}             & \emph{E+L} & 0.0429 &                          & 0.1904                           & 0.5062                           \\
ReStyle$_{LSAP}$      & \emph{E+L} & 0.0296 & \textcolor{red}{\textbf{-31.1\%}}                         & 0.1506                           & 0.6148 \\
HFGI$_{e4e}$ \cite{wang2022high}             & \emph{E+F} & 0.0296    &                       & 0.1172                           & 0.6816                           \\
HFGI$_{LSAP}$      & \emph{E+F} & 0.0210  &\textcolor{red}{\textbf{-29.0\%}}                         & 0.0945                           & 0.7405 \\
SAM$_{e4e}$ \cite{parmar2022spatially}              & \emph{E+F} & 0.0143  &                         & 0.1104                           & 0.5568                           \\
SAM$_{LSAP}$        & \emph{E+F} & 0.0117 &\textcolor{red}{\textbf{-18.1\%}} & 0.0939 & 0.6184                           \\
PTI$_{e4e}$ \cite{roich2021pivotal}             & \emph{E+T} & 0.0074 & & 0.0750 & 0.8633                           \\
PTI$_{LSAP}$             & \emph{E+T} & \textbf{0.0067} & \textcolor{red}{\textbf{-9.4\%}} & \textbf{0.0666} & \textbf{0.8696}   \\
\bottomrule
\end{tabular}
}
\caption{\textbf{Fidelity results on face domain.}Reconstruction performance is reported for encoder-based (\emph{E}), optimization-based (\emph{O}), and two-stage methods, including latent codes refinement (\emph{E+L}), feature prediction (\emph{E+F}), and generator tuning (\emph{E+T}). Gain denotes the MSE improvement achieved by our method over the corresponding $X$+e4e baselines.}
\label{table:face_inversion}
\end{table*}

For the other domains, we compare LSAP$_E$ with the widely used encoder e4e to demonstrate the generality of our approach. As shown in Table~\ref{table:other_domain}, LSAP$_E$ consistently achieves better performance across all three domains, indicating that our alignment strategy is robust and broadly effective for GAN inversion tasks.

Moreover, we assess perception and editability using NSCD, LEC and identity preservation during manipulation, as reported in Table~\ref{table:face_perception_edit}. LSAP$_E$ achieves the best NSCD and LEC across all three editing attributes, and attains the highest identity similarity in two of them. It is noteworthy that, although e4e provides reasonable editability, it exhibits poorer identity preservation than pSp under the "pose" and "smile" edits, largely due to its reconstruction gap. In contrast, LSAP$_E$ achieves higher similarity under the "pose" and "age" edits, indicating that our approach is more effective at preserving portrait identity during manipulation.

\begin{table}[htbp]
\centering
\resizebox{0.46\textwidth}{!}
{
\begin{tabular}{llccc}
\toprule
\textbf{Domain}      & \textbf{Method} & \textbf{MSE} $\downarrow$ & \textbf{LPIPS} $\downarrow$ & \textbf{NSCD} $\downarrow$ \\ \hline
\multirow{2}{*}{\textbf{Car}}         & e4e    &  0.1201   &  0.3252     &  0.0646    \\
            & LSAP$_E$   &   \textbf{0.1049}  &   \textbf{0.3106}    &  \textbf{0.0492}    \\ \midrule
\multirow{2}{*}{\textbf{Church}}      & e4e    &  0.1505   &   0.4307    &  0.0761    \\
            & LSAP$_E$   &  \textbf{0.1144}   &  \textbf{0.3426}     & \textbf{0.0588}     \\ \hline
\multirow{2}{*}{\textbf{Wild Animal}} & e4e    & 0.0882$^\dagger$    & 0.2658$^\dagger$      & 0.0379$^\dagger$     \\
            & LSAP$_E$   & \textbf{0.0785}    &   \textbf{0.2524} & \textbf{0.0224} \\
\bottomrule
\end{tabular}
}
\caption{\textbf{Quantitative results on other domains.} The symbol $^\dagger$ indicates that the original model is unavailable and the encoder is trained using the official code.}
\label{table:other_domain}
\end{table}

\begin{table}[htbp]
\centering
\resizebox{0.46\textwidth}{!}
{
\begin{tabular}{lcccc}
\toprule
\multirow{2}{*}{\textbf{Method}} & \multirow{2}{*}{\textbf{NSCD}}             & \multicolumn{3}{c}{\textbf{LEC / Similarity}}      \\ \cmidrule(lr){3-5}
    &            & \textbf{Pose}     & \textbf{Smile}        & \textbf{Age}\\ \midrule
pSp      & 0.10     & 89.35/0.43     & 55.86/\textbf{0.47}   & 64.61/0.30\\
e4e      & 0.04 & 26.65/0.41 & 22.32/0.41 & 23.28/0.34\\
LSAP$_E$ & \textbf{0.03} & \textbf{19.02}/\textbf{0.45} & \textbf{14.03}/0.45 & \textbf{14.67}/\textbf{0.39}\\
\bottomrule
\end{tabular}
}
\caption{\textbf{Perception and editability results on the face domain.} We report NSCD, LEC \cite{tov2021designing}, and identity similarity \cite{deng2019arcface} for three encoder-based inversion methods.}
\label{table:face_perception_edit}
\end{table}

\subsection{Qualitative Results}

\begin{figure*}[htbp]
	\centering
	\includegraphics[width=1.0 \linewidth]{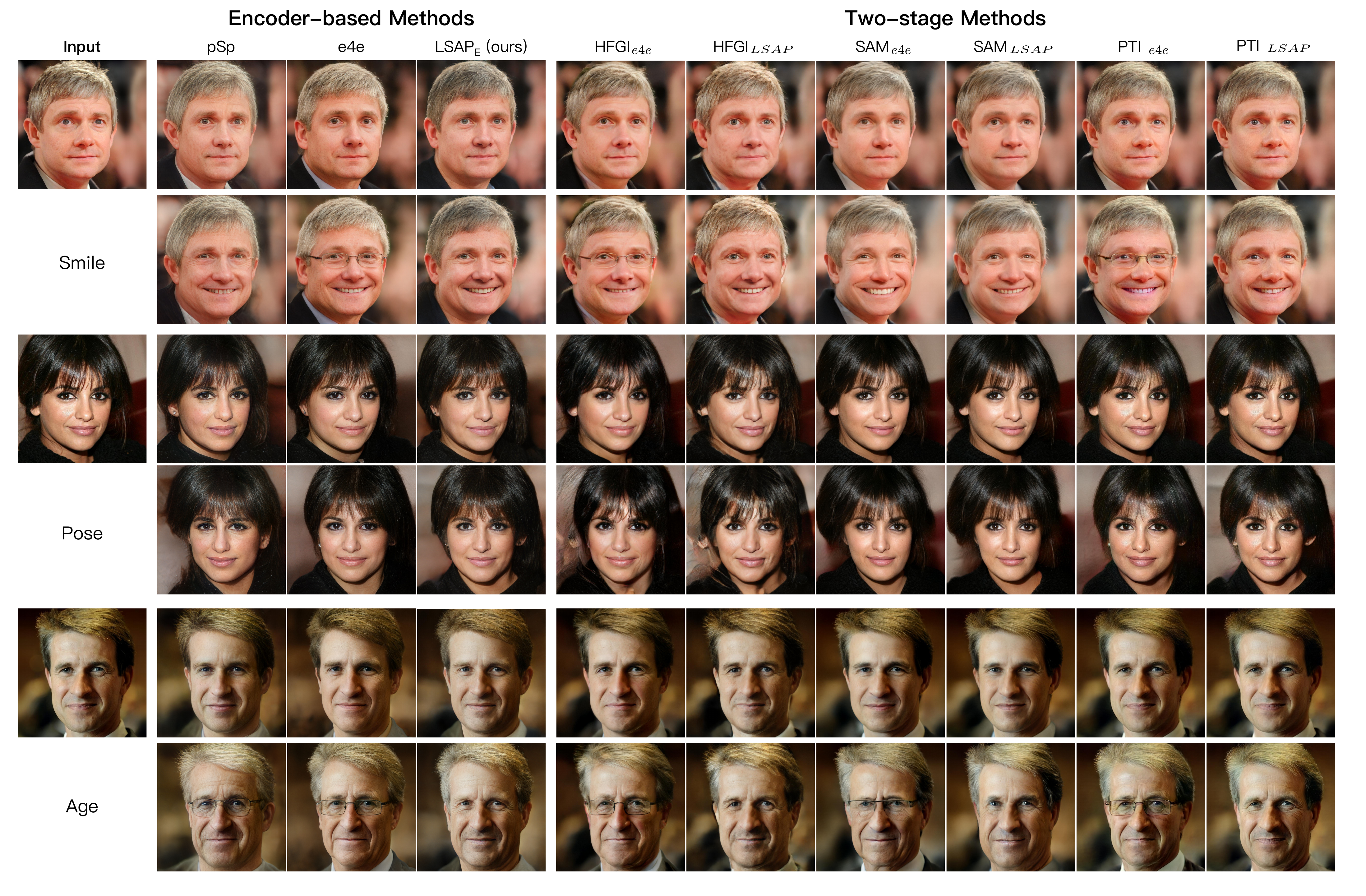}
	\caption{\textbf{Inversion and editing results of encoder-based and two-stage inversion methods on face domain.} We compare encoder-based, optimization-based, and two-stage approaches. LSAP$_E$ enhances perception and editability while maintaining fidelity, and HFGI$_{LSAP}$, SAM$_{LSAP}$ and PTI$_{LSAP}$ further reduce image distortion.}
	\label{fig:inversion}
\end{figure*}

We present qualitative comparisons in Figure~\ref{fig:inversion}.  In terms of reconstruction, our alignment paradigm achieves reconstruction quality comparable to pSp,  while substantially enhancing image perception and editability. Compared with e4e, LSAP$_E$ yields better fidelity and stronger editing performance. For instance, in the first example of Figure~\ref{fig:inversion} the editing result produced by e4e introduces redundant glasses under the "smile" edit. Across the two-stage methods, HFGI, SAM and PTI improve the reconstruction quality for both e4e and LSAP$_E$. Their inversion outputs and editing effects remain consistent with those from the corresponding encoders, while better retaining fine image details. Among them, PTI$_{LSAP}$ achieves state-of-the-art performance, offering the best combination of fidelity, perception quality, and editability.

\begin{figure*}[htbp]
	\centering
	\includegraphics[width=1.0 \linewidth]{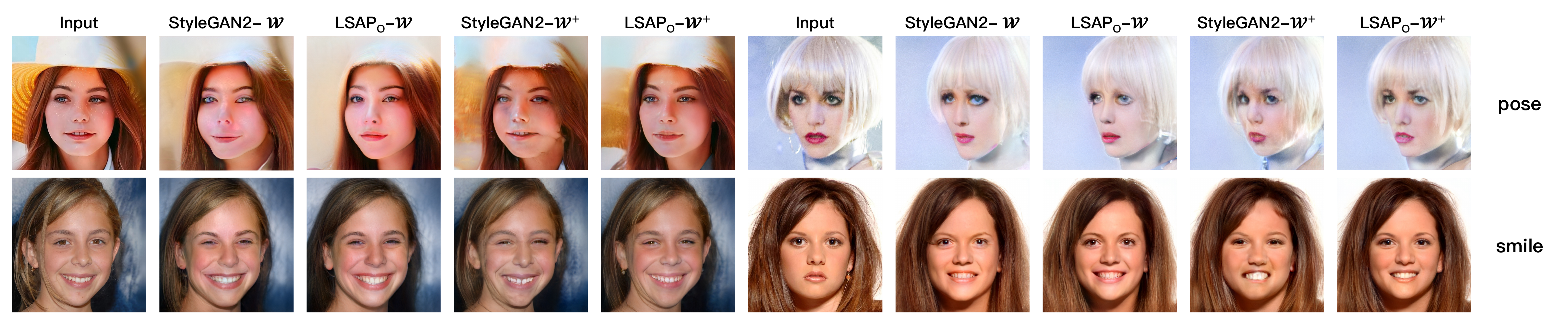}
	\caption{\textbf{Editability effects of LSAP for optimization-based methods.} LSAP enhances the editability of optimized latent codes and improves image quality in both the $\mathcal{W}$ and $\mathcal{W}^+$ spaces.}
	\label{fig:optim_edit}
\end{figure*}

For optimization-based methods, our approach enables the optimized latent codes to remain editable, as illustrated in Figure~\ref{fig:optim_edit}. Vanilla projection in both the $\mathcal{W}$ and $\mathcal{W}^+$ spaces tends to produce unnatural facial details, whereas applying LSAP leads to substantial improvements. This demonstrates that our method offers a concise and effective solution even for optimization-based inversion strategies.

\begin{figure*}[htbp]
	\centering
	\includegraphics[width=1.0 \linewidth]{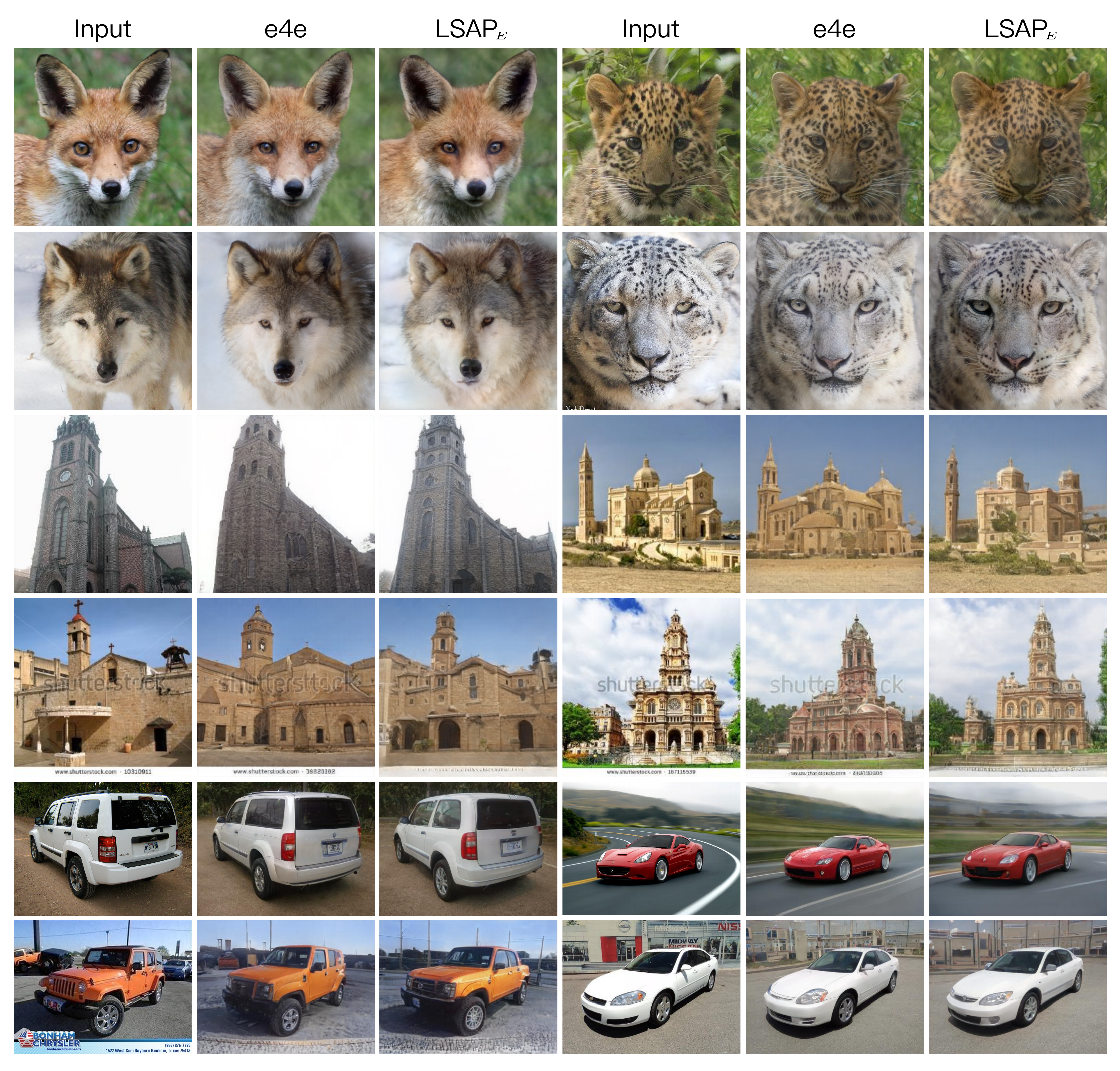}
	\caption{\textbf{Inversion results on other domains.} For the car and church domains, official e4e models are available, while for the wild animal domain, we train the encoder on the AFHQ-Wild dataset \cite{choi2020stargan}. }
\label{fig:other_domains}
\end{figure*}

\begin{figure}[htbp]
	\centering
	\includegraphics[width=0.6 \linewidth]{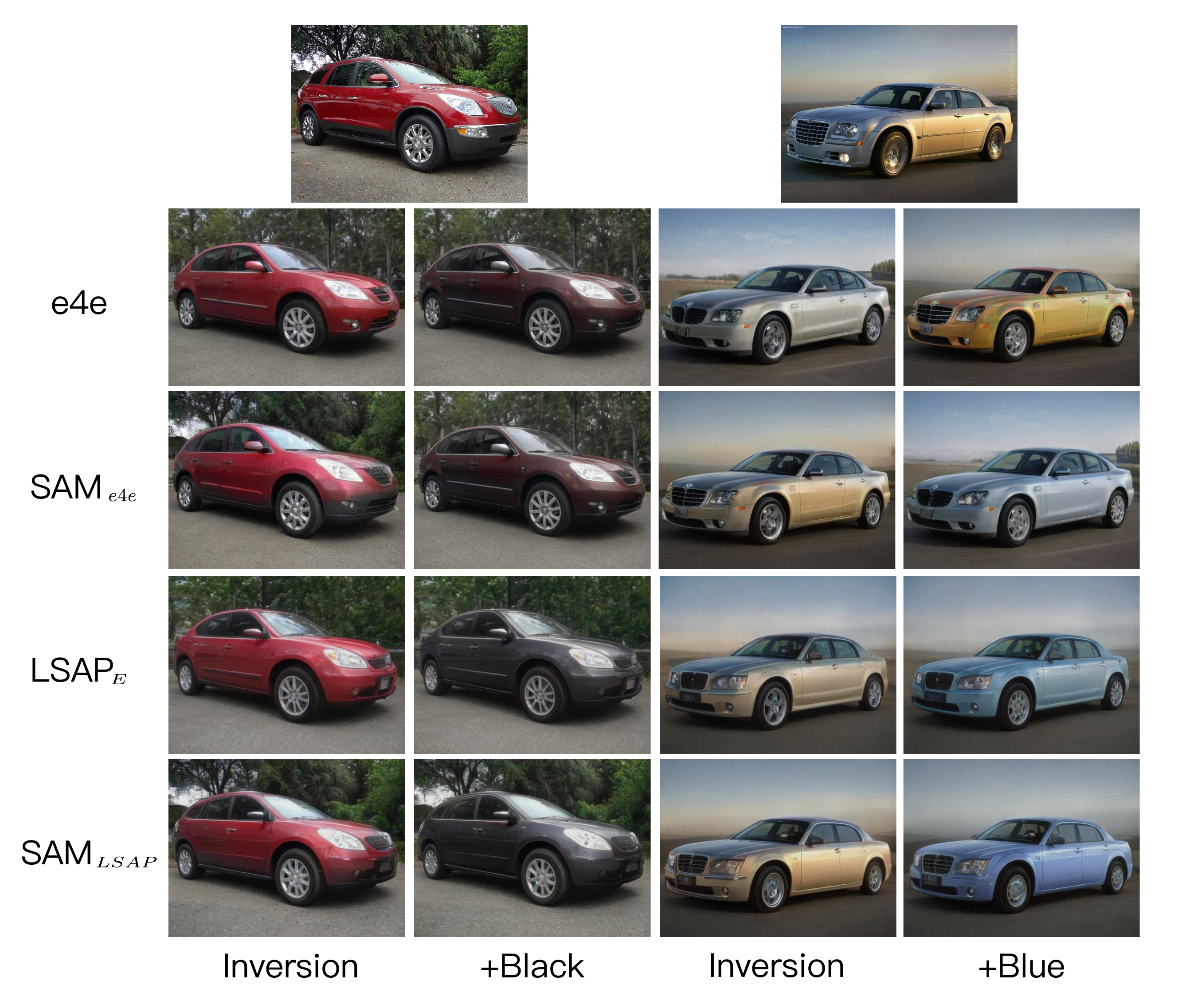}
	\caption{\textbf{Inversion and editing comparison between e4e and LSAP$_E$.} We show the inversion and editing results of both encoders, along with their corresponding outputs using SAM. LSAP substantially improves editability and preserves more visual details during inversion.}
	\label{fig:car_edit}
\end{figure}

We further visualize the inversion and editing results of e4e and LSAP$_E$ in other domains, including cars, churches, and wild animals. The results are shown in Figure~\ref{fig:other_domains} and Figure~\ref{fig:car_edit}. In terms of inversion, LSAP$_E$ provides a slight fidelity improvement, particularly in reconstructing color and reflections more accurately. For instance, in the second example of Figure~\ref{fig:car_edit}, faithfully preserved in the LSAP$_E$ result,whereas the output from e4e exhibits only a flat white region. During editing, LSAP$_E$ strong capability in generating high-quality manipulation results. When combined with the SAM technique,  LSAP$_E$ achieves superior performance in both inversion and editing across these domains.

\subsection{Perception and Editability in Two Stages} In \S~\ref{sec:inversion_ability}, we note that fidelity is mainly improved during the \emph{Result Refinement} stage, whereas perception and editability are largely determined by the \emph{Image Embedding} step. As illustrated in Figure~\ref{fig:inversion}, when the encoder produces weak editing results—such as attribute entanglement—the outputs of the two-stage methods remain similar. For example, when editing the third image with the "age" attribute, e4e introduces unintended glasses, and the corresponding results from HFGI$_{e4e}$, SAM$_{e4e}$, and PTI$_{e4e}$ exhibit the same artifact. Thus, although \emph{Result Refinement} significantly enhances fidelity, the \emph{Image Embedding} step remains crucial in the inversion pipeline.

\subsection{Image Perception}
\label{sec:appendix_image_perception}

\begin{figure*}[htbp]
	\centering
	\includegraphics[width=0.8 \linewidth]{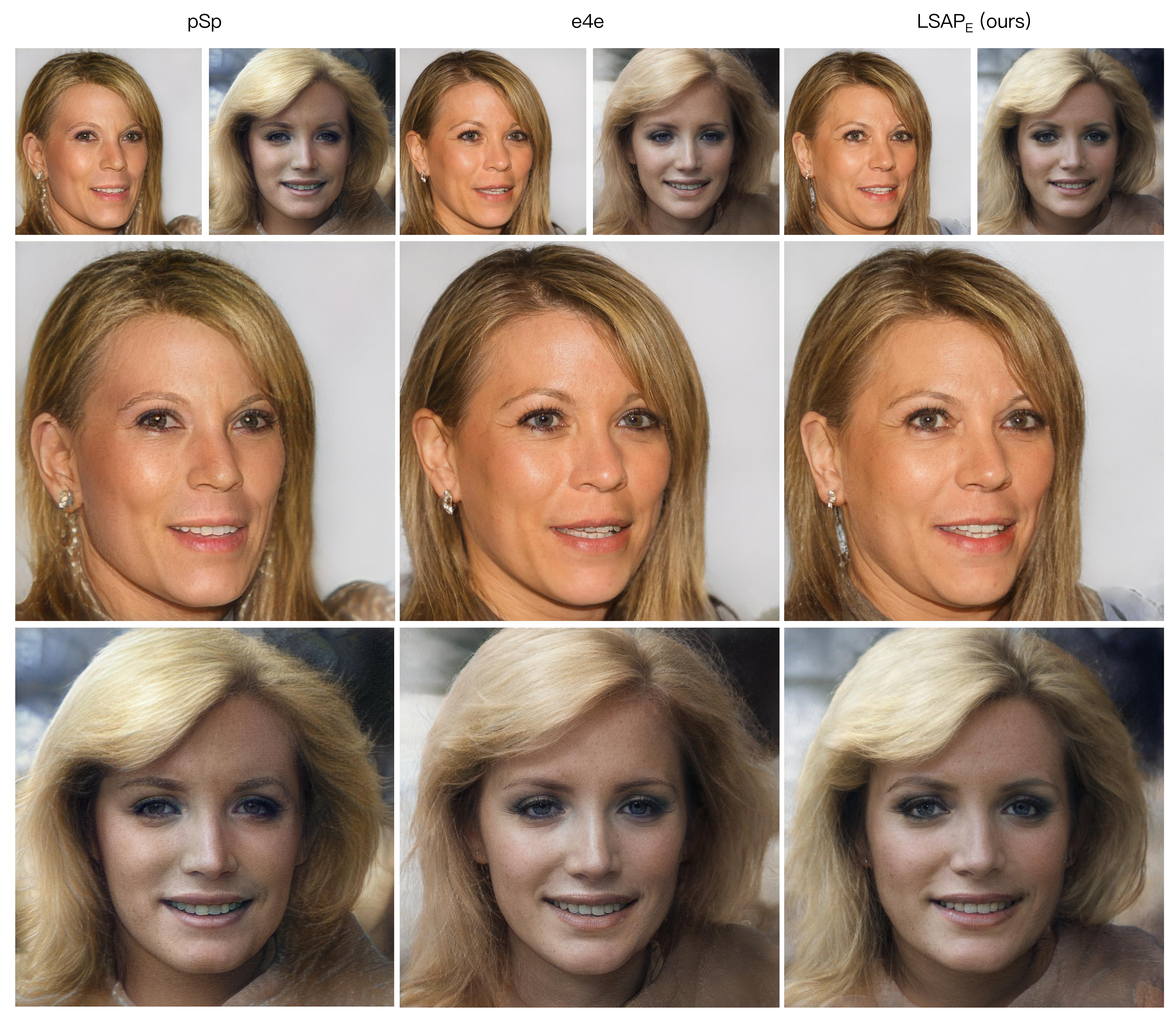}
	\caption{\textbf{High-resolution visualization of image perception}. We present high-resolution inversion results from pSp, e4e, and LSAP$_E$ to highlight fine image details, with corresponding low-resolution results shown in the first row for comparison. Differences in perceptual quality are less noticeable at low resolution.}
	\label{fig:image_perception}
\end{figure*}

We illustrate the differences in image perception among inversion methods using high-resolution reconstruction results. As shown in Figure~\ref{fig:image_perception}, the inverse outputs from each approach exhibit subtle but noticeable variations at high resolution, particularly in regions such as hair, teeth, lips, and skin. These discrepancies are less apparent at lower resolutions or in thumbnail views, as demonstrated in the first row, but they become much more evident—and often lead to unnatural or artificial appearances—when viewed at higher resolution. Therefore, we recommend evaluating visual quality at resolutions such as (e.g., $1024\times 1024$). Our alignment paradigm effectively improves perceptual quality; as shown in Figure~\ref{fig:image_perception}, the reconstructed images produced by our method retain more natural and realistic fine details.

\subsection{NSCD} In \S~\ref{sec:disalignment} we show that these two characteristics are closely related to the alignment between the inverse codes and the synthetic distribution. We then introduce NSCD as a numerical measure of this alignment, which is validated by our experiments. Methods with smaller NSCD values exhibit higher image quality and stronger editing performance (e.g., e4e and LSAP$_E$), whereas larger NSCD values indicate weaker reconstruction and manipulation capability (e.g., pSp and StyleGAN2-$\mathcal{W}^+$). Compared with LEC \cite{tov2021designing}, NSCD is computed directly on latent codes and is independent of the editing direction, making it a more general and convenient metric.

\subsection{Ablation Study}
\label{sec:appendix_ablation}

\begin{figure*}[htpb]
	\centering
	\includegraphics[width=0.5 \linewidth]{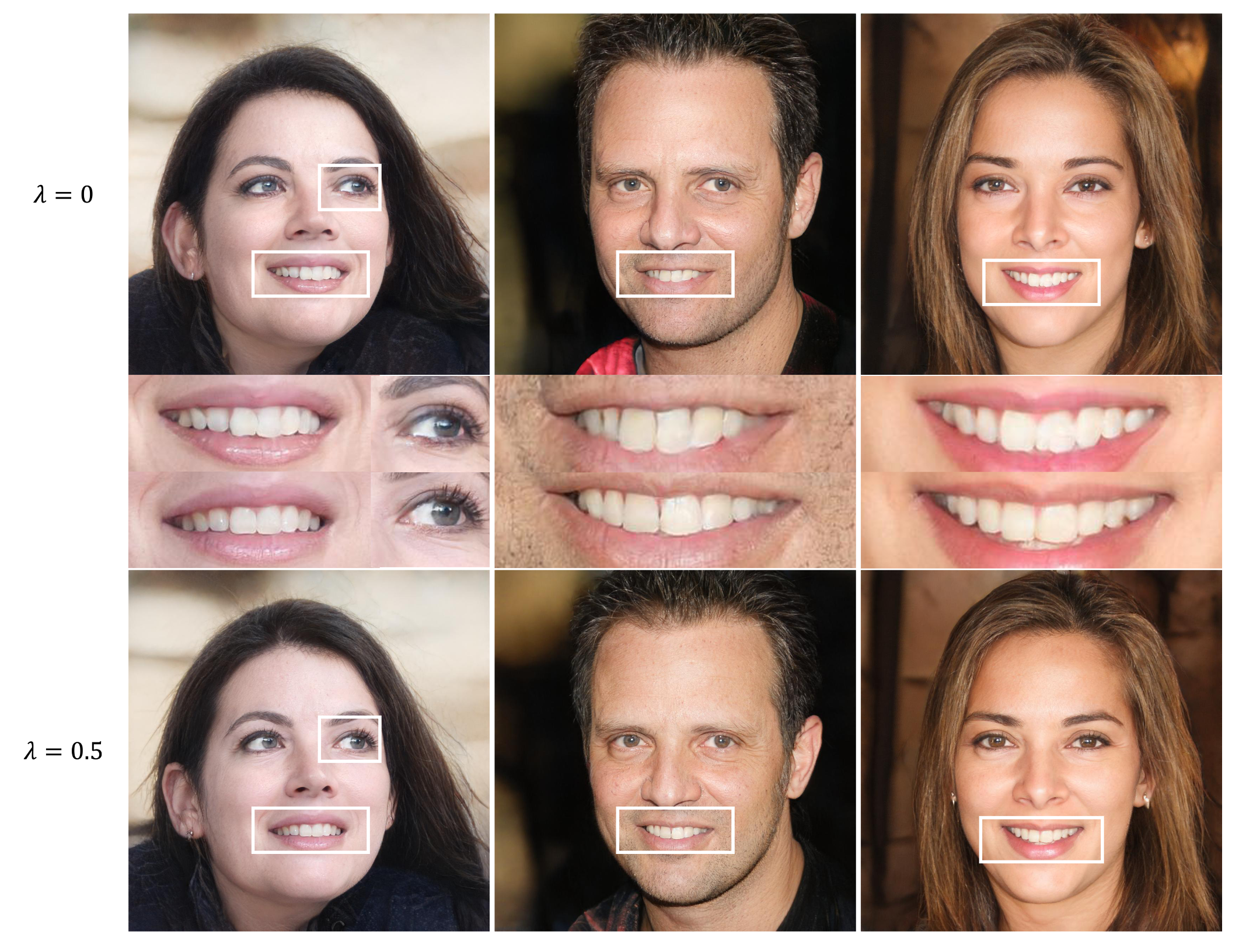}
	\caption{\textbf{Ablation study on image perception.} We compare inversion results from LSAP$_E$ with those from the same encoder without $\mathcal{L}_{NSCD}$ to illustrate its effect. LSAP$_E$ substantially improves image quality and eliminates unnatural artifacts.}
	\label{fig:ablation}
\end{figure*}

\begin{figure*}[htpb]
	\centering
	\includegraphics[width=0.5 \linewidth]{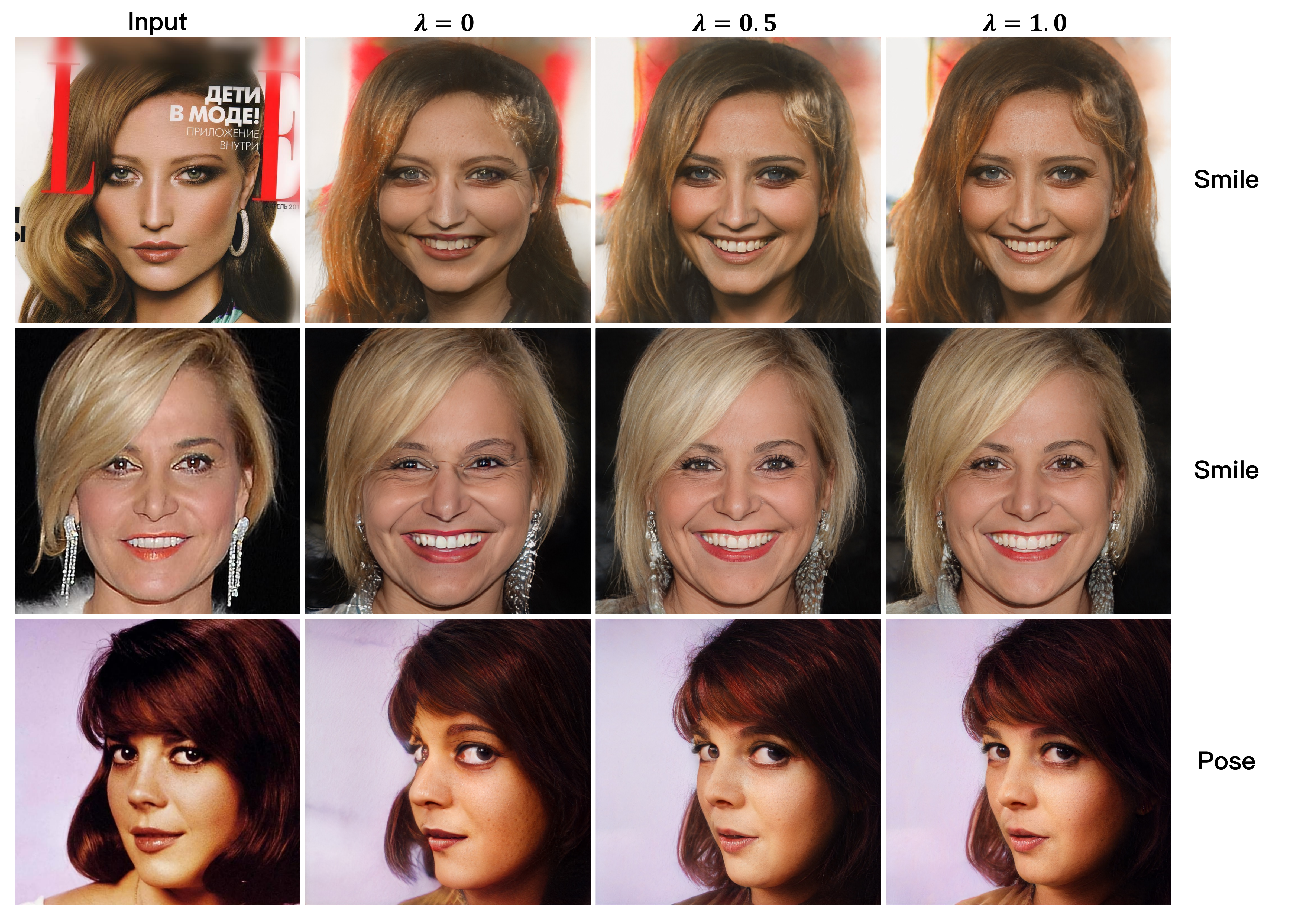}
	\caption{\textbf{Ablation study on image editability.}We present manipulation results from LSAP$_E$ under different values of the hyper-parameter $\lambda$.}
	\label{fig:ablation_edit}
\end{figure*}

We examine the hyper-parameter $\lambda$ of $\mathcal{L}_{NSCD}$ on the face domain using  LSAP$_E$ as an example, and the quantitative results are reported in Table ~\ref{table:ablation}. Increasing $\lambda$ leads to higher image distortion, which aligns with our expectation since $\lambda$ controls the strength of the alignment loss. Conversely, perception and editability improve as $\lambda$ increases. Figure~\ref{fig:ablation} visualizes inversion results for $\lambda=0$ and $\lambda=0.5$. In the first row ($\lambda=0$), the textures of the teeth, eyes, and lips are noticeably degraded. For instance, in the left image of the first row, the end of the left eyelid (right side in the figure) is positioned unrealistically far from the eye. In addition, the teeth exhibit misalignment and sticking artifacts, and the lips appear overly smooth and lack natural texture. These issues are effectively resolved with LSAP, as shown in the second row. To illustrate the change in editability, we further compare manipulation results under $\lambda=0,0.5$ and $1.0$, as shown in Figure~\ref{fig:ablation_edit}. The first two examples correspond to the "smile" edit and the third to the "pose" edit. When $\lambda=0$, the edited images are less photorealistic, and unwanted glasses appear under the "smile" edit. In contrast, the results for $\lambda=0.5$ and $1.0$ are visually similar and exhibit strong editability. Together, the inversion and editing results highlight the effectiveness and superiority of our alignment paradigm.

\begin{table*}[htbp]
\centering
\resizebox{0.7\textwidth}{!}
{
\begin{tabular}{lccccccc}
\toprule

\multirow{2}{*}{\textbf{$\lambda$}}      & \multicolumn{3}{c}{\textbf{Fidelity}}                                                                         & \multicolumn{4}{c}{\textbf{Perception \& editability}}                                                                                                                                                                     \\
\cmidrule(r){2-4} \cmidrule(r){5-8}
                    & \textbf{MSE} $\downarrow$                 & \textbf{LPIPS} $\downarrow$               & \textbf{Similarity} $\uparrow$            & \textbf{NSCD} $\downarrow$                & \textbf{LEC}$_{pose}$ $\downarrow$         & \textbf{LEC}$_{smile}$ $\downarrow$ & \textbf{LEC}$_{age}$ $\downarrow$ \\ \hline
0            & 0.0369 & 0.1657 & 0.5512 & 0.0736  & 24.8245  & 22.5007 & 24.8069 \\
0.1            & 0.0382 & 0.1703 & 0.5438 & 0.0416  & 19.1594  & 14.0133 & 15.2246\\
0.25           & 0.0391 & 0.1737 & 0.5410 & 0.0395  & 19.1345  & 14.1382 & 15.1599\\  
\textbf{0.5}   & 0.0397 & 0.1766 & 0.5305 & 0.0385 & 19.0211 & 14.0360 & 14.6715\\ 
0.75           & 0.0406 & 0.1792 & 0.5222 & 0.0381  & 19.0949  & 14.0128 & 14.3198\\ 
1.0            & 0.0413 & 0.1809 & 0.5168 & 0.0378  & 15.8013  & 13.8433 & 14.6084\\
\bottomrule
\end{tabular}
}
\caption{\textbf{Ablation study on the hyper-parameter of LSAP$_E$.} We use $\lambda=0.5$ as the default setting in our experiments.}
\label{table:ablation}
\end{table*}

\section{Conclusion}
Fidelity, perception, and editability are three essential characteristics of GAN inversion methods. We begin by analyzing the origins of these properties in the inversion pipeline and find that aligning the embedded images with the synthetic distribution during the Image Embedding stage is crucial—both for achieving high-fidelity reconstructions and for improving the subsequent Result Refinement process.Motivated by this observation, we introduce the Latent Space Alignment Inversion Paradigm (LSAP), which addresses the measurement and correction of latent space disalignment. To quantitatively and intuitively characterize this misalignment, we propose Normalized Style Space Cosine Similarity (NSCD) as a metric, defined in the Normalized Style Space ($\mathcal{S^N}$) Owing to its differentiable formulation, NSCD enables a unified alignment Extensive experiments across four domains and three categories of baselines demonstrate that LSAP consistently improves fidelity, perception, and editability. Moreover, two-stage inversion methods combined with LSAP achieve state-of-the-art performance.



\bibliography{egbib}

\end{document}